# Symposium on Data Science and Statistics (SDSS 2024)
# Bias Correction in Machine Learning-based Classification of Rare Events

**Luuk Gubbels**  **Marco Puts**  **Piet Daas**

### Abstract

Online platform businesses can be identified by using web-scraped texts. This is a classification problem that combines elements of natural language processing and rare event detection. Because online platforms are rare, accurately identifying them with Machine Learning algorithms is challenging. Here, we describe the development of a Machine Learning-based text classification approach that reduces the number of false positives as much as possible. It greatly reduces the bias in the estimates obtained by using calibrated probabilities and ensembles.

**Keywords:** Calibration, Population, Ensembles

## 1 Introduction

Obtaining reliable information from a small or rare subpopulation is a challenging topic for many researchers. Approaches commonly used to find rare or so-called hard-to-identify groups are a screening survey, network sampling, area sampling, or a combination (Snijkers et al. 2013). Examples of a rare subpopulation are online platforms. These are defined as:

> A digital service that facilitates interactions between two or more distinct but interdependent sets of users (whether firms or individuals) who interact through the service via the Internet (OECD, 2019).

To obtain a complete overview of all online platforms, a Support Vector Machine (SVM) classification model was developed and applied to the entire population of businesses with a website in the Netherlands. Here, it was found that: i) online platforms compose of 0.22% of the total population of businesses with a website and ii) considerable manual checking was needed in this process (Daas et al., 2023). The latter was predominantly the result of the large number of false positives produced by the model. In this paper, we describe the development of a fully automated approach that aims to reduce the number of false positive online platforms detected as much as possible. This, consequently, seriously reduces the bias in the total estimate of the number of online platforms obtained and the manual checking needed.

## 2 Data and methods

All websites assigned to businesses in the Business Register (BR) of Statistics Netherlands were studied. This resulted in a list of 960,588 (unique) URLs of which 629,284 could actually be scraped (Daas et al., 2023). Web scraping was performed as described in Daas and van der Doef (2021) and up to a maximum of 200 pages was collected per URL. The texts were extracted and processed as described in Daas et al. (2023). Per URL, the texts were combined.

For online platform classification, a set of 569 online platforms (positives) were identified by experts. To this set, a random sample of 1328 non-platforms (negatives), from the scraped websites linked to the BR, were added. This resulted in an 1897-sized dataset with 30% platforms and 70% non-platforms. To ensure the independent evaluation of the findings, a 228-sized dataset was created, containing 69 positive and 159 negative cases, by sampling (and removing) them from the 1897-sized dataset. The resulting 1669 dataset was used for model development.

All scripts were written in Python (v.3.7) and the Machine Learning algorithms implemented in scikit-learn (v.0.24) were used. The Bayesian calibration method of Puts and Daas (2021) was used to correct the intrinsic prevalence of probability-producing classification models. The code is available on GitHub (Puts, 2023). Multiple algorithms were tested but in this paper, only the results of Logistic Regression based models are shown. This classification model provided results almost similar to that of the original SVM model but could be trained much faster (Gubbels, 2023). Multiple models, up to 10, were trained on random resamples (bootstraps) of the 1669-sized training set. The bootstraps were randomly split into a 70% training and a 30% test set. A 10% sample of each training set was used as a validation set. The accuracy of the trained model on the validation set was used to obtain the weight used for voting. The results of classifying website texts with multiple models (ensembles of up to 10 models) were obtained by using the average of the weighted votes. More details are described in Gubbels (2023).

## 3 Results

The downside of the original model developed was the considerable number of false positive online platforms detected (Daas et al., 2023). This resulted in an overestimation of the number of platforms, i.e. producing a bias in the estimate (Table 1, row 1). Extensive manual checking needed to be performed to deal with this issue. Below, various steps are described to considerably reduce the number of false positives.

### 3.1 Using probabilities

Apart from producing binary labels, a Logistic

Regression model can also produce the probability of a website being an online platform. These probabilities have a value between 0 and 1 and depend on certain assumptions to be considered actual probabilities; such as a normal distribution of these values (Gubbels, 2023). Using the 'probabilities' of the model was found to increase the online platform estimates; hence, increasing the bias (Table 1, row 2). This makes clear that, under these circumstances, the model did not produce actual probabilities. This did not affect the accuracy (Acc.) and balanced accuracy (B. Acc.).

|   | Type of model | TP | Est.Pos. | Bias | Acc. | B. Acc. |
|---|---|---|---|---|---|---|
| 1. | Log. Reg | 69 | 2991 | 0.098 | 0.901 | 0.0907 |
| 2. | Log. Reg, prob. | 69 | 7657 | 0.255 | 0.901 | 0.0907 |
| 3. | Log. Reg, cal. prob. | 69 | 637 | 0.019 | 0.985 | 0.7610 |
| 4. | Ensemble, cal. prob. | 69 | 306 | 0.007 | 0.993 | 0.6275 |

Table 1: Effect of various model-based approaches applied.

### 3.2 Calibrating probabilities

Subsequently, a method was applied to correctly calibrate the probabilities of the Logistic Regression model. This method corrects for the intrinsic prevalence of the model; i.e. the prevalence caused by the ratio of positives and negatives on which the model was originally trained (Puts and Daas, 2021). Since the number of online platforms is very low in the population and the model was trained on 30% positive and 70% negative cases, it can be expected that correcting for this prevalence may seriously reduce the number of positive cases estimated. Applying the calibration method revealed that this was indeed the case; see Table 1, row 3. As a consequence, both the accuracy and the balanced accuracy increased considerably.

### 3.3 Using an Ensemble

When the results of multiple trained and calibrated models, up to 10, were combined, the bias was reduced even further; see Table 1, row 4. The accuracy increased but the balanced accuracy reduced somewhat. The latter was likely the result of optimizing the model-based findings on the 30% positive and 70% negative cases in the dataset used, which must have negatively affected the balanced accuracy calculated on 50% of each group. Figure 1 graphically illustrates the effect on the bias of the various model-based approaches applied.

## 4 Discussion

From the results in Table 1 and Figure 1, it is clear that the classification findings of a Logistic Regression model can be improved by calibrating its probabilities and combining the findings of multiple calibrated models. The calibration step improves the estimate by greatly reducing the number of false positives produced. This seriously diminishes the number of websites that need to be checked manually. The ensemble approach improves the estimate even more by combining the findings of multiple calibrated models, each trained on slightly different datasets. However, the bias is not completely removed and the number of (false) positives is still considerable (Table 1, row 4). This is not unexpected for a model detecting rare events. We believe there are two ways to even further improve this approach. The first is by increasing the number of models included in the ensemble. The second is reducing the (potential) difference in the composition (the representativeness) of the websites included in the dataset used to train the model(s) and those in the dataset used to produce the findings listed in Table 1. Both are topics that will be studied in future research.

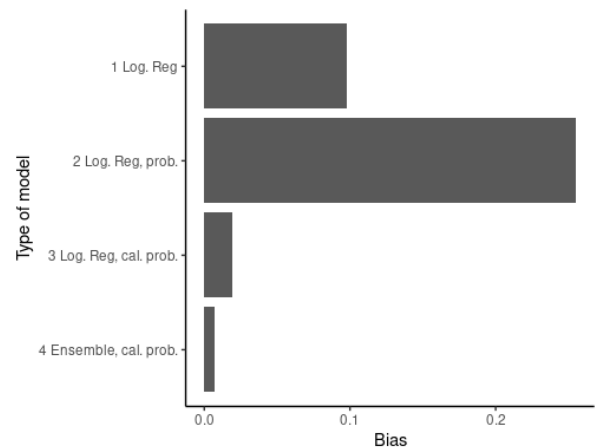

Figure 1: Effect on the bias of the model-based approaches.